\documentclass{article}

\usepackage{arxiv}

\usepackage[utf8]{inputenc} 
\usepackage[T1]{fontenc}    
\usepackage{hyperref}       
\usepackage{url}            
\usepackage{booktabs}       
\usepackage{amsfonts}       
\usepackage{nicefrac}       
\usepackage{microtype}      
\usepackage{graphicx}
\usepackage[ruled,vlined]{algorithm2e}
\usepackage{algorithmic}
\usepackage{breakurl}
\usepackage{lineno}
\usepackage{subfig}
\title{Efficient Retail Video Annotation: A Robust Key Frame Generation Approach for Product and Customer Interaction Analysis}

\author{%
  Varun Mannam\thanks{Corresponding author: Dr. Varun Mannam is an Applied Scientist II currently in People eXperience Technology (PXT) Central Science at Amazon, where he leads research initiatives in talent analytics and intelligent HR systems. Disclosure: This research was conducted at Alexa (AGI) within Amazon Data Services (ADS)-Science during 2022-2023.} \\
  Alexa, AGI (ADS-Science) \\
  Amazon\\
  Seattle, WA 98004 \\
  \texttt{mannamvs@amazon.com} \\
  \And
  Zhenyu Shi \\ 
  Alexa, AGI (ADS-Science) \\
  Amazon\\
  Seattle, WA 98004 \\
  \texttt{szhenyu@amazon.com} \\
}
\date{TODAY}

\begin{document}
\vspace{-1cm}
\maketitle
\vspace{-1cm}
\begin{abstract}
Accurate video annotation plays a vital role in modern retail applications, including customer behavior analysis, product interaction detection, and in-store activity recognition. However, conventional annotation methods heavily rely on time-consuming manual labeling by human annotators, introducing non-robust frame selection and increasing operational costs. To address these challenges in the retail domain, we propose a deep learning-based approach that automates key-frame identification in retail videos and provides automatic annotations of products and customers. Our method leverages deep neural networks to learn discriminative features by embedding video frames and incorporating object detection-based techniques tailored for retail environments. Experimental results showcase the superiority of our approach over traditional methods, achieving accuracy comparable to human annotator labeling while enhancing the overall efficiency of retail video annotation. Remarkably, our approach leads to an average of 2-times cost savings in video annotation. By allowing human annotators to verify/adjust less than 5\% of detected frames in the video dataset, while automating the annotation process for the remaining frames without reducing annotation quality, retailers can significantly reduce operational costs. The automation of key-frame detection enables substantial time and effort savings in retail video labeling tasks, proving highly valuable for diverse retail applications such as shopper journey analysis, product interaction detection, and in-store security monitoring.
\end{abstract}
\vspace{-0.2cm}

\section{Introduction}
In recent years, computer vision models have made remarkable advancements in various domains, including image recognition \cite{russakovsky2015imagenet}, object detection \cite{lin2014microsoft}, object tracking \cite{wojke2017simple}, semantic image segmentation \cite{everingham2015pascal}, video summarization \cite{otani2017video} and video recognition \cite{chen2019drop, arnab2021vivit}. These models have achieved outstanding performance on challenging benchmarks, thanks to the availability of large annotated datasets. However, acquiring accurate labels for these datasets is often time-consuming and expensive - a challenge particularly acute in the retail industry where rapid product turnover and diverse customer interactions demand continuous annotation updates. Despite recent progress in self-supervised learning \cite{sermanet2018time} and weak supervised learning \cite{deng2019accurate}, deep learning models for computer vision still heavily rely on high-quality labeled data. While transfer learning \cite{touvron2021training}, fine-tuning \cite{cui2018large, li2020rethinking}, and few-shot learning \cite{finn2017model} methods have attempted to reduce this dependency \cite{wang2020generalizing}, they still require expensive and high-quality labels during training. This issue is particularly pronounced in retail video annotation, which presents unique challenges such as the need to recognize visually similar products, track multiple shoppers simultaneously, and identify subtle customer-product interactions.
\begin{figure*}[!t]
  \centering
  \includegraphics[width=0.95\linewidth]{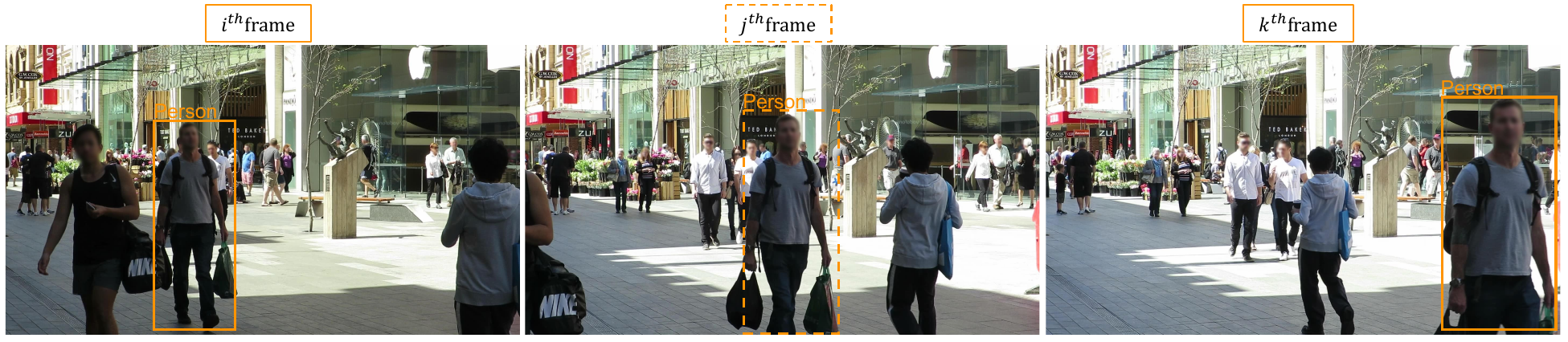}
  \caption{Overview of the current video annotation system. To reduce the Average Annotation Time (AAT), the human annotator labels a down-sampled version of the video \protect\footnote{\url{https://motchallenge.net/sequenceVideos/ADL-Rundle-3-raw.webm}}, annotating only select key-frames. For example, the human annotator annotates only the $i^{th}$ and $k^{th}$ frames as key-frames, and the $j^{th}$ frame can be automatically annotated using the interpolation method. The human annotations and auto annotations are shown in the solid and dashed lines, respectively. The proposed KFG method further optimizes this process by identifying the minimum number of key-frames requiring human verification.}
  \label{current-labeling}
\end{figure*}

Recently, researchers have been investigating automated solutions \cite{muller2018trackingnet, vondrick2011video, manen2017pathtrack} to streamline the process of video annotation. A fundamental approach involves soliciting human annotators to label bounding boxes for a subset of frames, while leveraging temporal interpolation or advanced time-variation detection algorithms to generate annotations for the remaining frames. In the retail domain, video annotation requirements span multiple applications: in-store customer journey analysis, product interaction detection, inventory monitoring, and theft prevention systems. The annotation data is used to train deep learning models for detecting customers, products, staff members, and their interactions. Retail video annotation often involves multiple intertwined tasks at different granularities. With a single retail video, it is common to encounter at least three distinct annotation tasks. First, annotate store-wide attributes throughout the entire clip, such as store section or crowd density. Second, annotate objects, including bounding boxes for customers, products, and staff members. Last, per-frame annotation, focusing on customer-product interactions and shopping behaviors. While existing annotation pipelines incorporate automatic bounding box generation with manual adjustment capabilities, they do not guarantee annotation quality or cost efficiency. Hence, there is a need to address these limitations and explore methods that ensure both efficiency and high-quality annotation in retail environments.

Despite the efforts made by researchers to establish an annotation workflow, the inherent complexity of retail scenes often leads to various annotation errors and non-robust frame selection. Currently, there is no formal mechanism in place to detect and address robust frame selection in the video annotation tasks performed by human annotators. Figure.\ref{current-labeling} shows the typical video annotation process. One example of the user-selection introduced by the human annotator is the selective choice of frames to be annotated, which clearly introduces a non-robust method in the annotation workflow. To illustrate the user-selection of the frames, Figure.~\ref{current-labeling} shows the existing non-robust frame selection in the video annotation system. Here, the user can select any frame index as the key-frame and perform the frame-wise annotation (such as $i^{th}$ or $k^{th}$ frames shown in the solid-lines). However, this process is time-consuming and particularly inefficient for retailers who need to analyze thousands of hours of video footage. If the user selects random frames as key-frames, then the selection includes a non-robust solution (varies across multiple human annotators) that leads to expensive annotation costs that impact retail profit margins. In order to reduce annotation costs, retail videos are typically annotated at fixed intervals defined by either the decimation rate (1/30 or 1/60) or the frames per second rate (FPS). The remaining frames in the video can be annotated using a linear interpolation method. However, this solution, while reducing the cost of video annotation, often provides inaccurate video annotations, causing poor data quality that compromises critical retail analytics and decision-making.

In this paper, we focus on the video annotation challenges in the retail domain, though our approach is potentially adaptable to other multi-modal domains. This paper makes the following contributions:
\begin{itemize}
    \item Identified the pain points of high cost in the existing retail video annotation workflow and insufficient performance of established video annotation approaches for capturing customer-product interactions.
    \item Demonstrated a method to identify robust key frames in retail video annotation and presented the efficiency and quality of the complete video annotation compared to human video annotation datasets of in-store scenarios.
    \item Our proposed key frame generation (KFG) method for retail image and video annotation reduces annotation cost and time while ensuring robust key frame selection for critical customer behaviors and product interactions.
\end{itemize}
Note, the approach described in this paper is specifically designed for creating person bounding boxes in retail environments. However, the underlying principles of the KFG method can apply to other retail-specific labeling tasks and data types. For example, the KFG method could identify key frames for product interaction detection, shopper journey analysis, or automated checkout systems.

\section{Related Work}
Obtaining high-quality annotation is expensive, particularly in retail environments where product displays and customer behavior are highly variable. Traditionally, data service providers assign multiple annotators to reprocess the same work item if the model predicts errors, a labor-intensive process known as relabeling. However, achieving consensus is difficult, especially for retail video annotation where products may be partially occluded or handled by customers. The predominant method to reduce retail video annotation costs is to select only key frames or critical frames for human annotation, with the remaining frames annotated using interpolation. Annotators typically select frames at predetermined intervals, such as every 1/30th frame (assuming 30 FPS), every 1/60th frame (60 FPS), or at the video's FPS. This traditional video annotation approach is common in retail settings (Figure.\ref{block-diagram}(a)).

In the literature, methods for key frame selection fall into two categories: traditional methods and deep-learning based methods. However, these approaches remain time-consuming for retail video annotation with their selected frames, which is the current approach used in video annotation tools such as SMGT and Scale AI solutions deployed by major retailers. Several methods have been proposed to identify key frames in videos for multiple downstream tasks such as video summarization \cite{huang2019novel, ma2002user}, motion detection, activity detection \cite{gawande2020deep}, video content identification \cite{chatzigiorgaki2009real}, video recognition \cite{qi2018cnn, ainasoja2018keyframe}, and video captions \cite{singh2021efficient}. In retail specifically, these methods have been explored for tasks like customer journey analysis, product interaction detection, and theft prevention. However, these methods are not effective in providing quality video annotations with their proposed key frames detection methods. 

\subsection{Traditional methods} \label{td_methods}
Traditional key frame detection in retail video employs histogram analysis across RGB channels and frame difference thresholding, marking significant changes (such as product selection moments) as key frames. Retail video processing typically categorizes frames as I-frames (complete images), P-frames, and B-frames, leveraging I-frames as natural keyframes. Our experiments with in-store footage demonstrated effective key frame identification using both absolute frame difference calculations and histogram comparisons. Recent unsupervised clustering approaches \cite{yan2020self, tang2023deep, jadon2020unsupervised} further enhance detection by grouping similar retail scenes, automatically identifying new clusters when significant events occur (e.g., new customer arrivals or product interactions). Adaptive clustering methods \cite{man2022interested} have proven particularly effective for capturing dynamic customer-product interactions in variable retail environments. Comparative results of these methods on retail footage are presented in our analysis.

\subsection{Deep learning methods} \label{dl_methods}
Deep learning significantly advances retail video annotation through frame embedding and clustering techniques \cite{otani2017video, fajtl2019summarizing, elahi2022online}. By encoding retail footage using CNNs (ResNet, MobileNet \cite{deng2020review}, Inception \cite{szegedy2016rethinking}) into fixed-size latent vectors (e.g., 2048-dimensional with ResNet-50), we capture subtle customer-product interactions traditional methods miss. Our approach normalizes embeddings (zero mean, unit variance), applies PCA for dimensionality reduction, and measures cosine distances between frames, revealing clear delineation between distinct shopping behaviors. K-means clustering on these embeddings identifies natural groupings, with centroids serving as representative key-frames, reducing annotation workload substantially.

Recent object detection advances offer complementary capabilities. While SAM \cite{kirillov2023segment} provides powerful segmentation without class labels, and DETR \cite{carion2020end} delivers bounding boxes with classification, both require retail-specific fine-tuning for optimal performance in distinguishing between product categories and customer interactions. Our method combines these approaches to detect retail key-frames with robust bounding box annotations, significantly reducing Average Annotation Time while maintaining high annotation quality. Human annotators merely verify uncertain detections, focusing effort where algorithmic confidence is low. Comparative retail implementation results are presented in Section~\ref{results_discuss}.

\section{Key Frame Generation (KFG) Method} \label{kfg_methods}
\begin{figure}[!ht]
  \centering
  \includegraphics[width=0.97\linewidth]{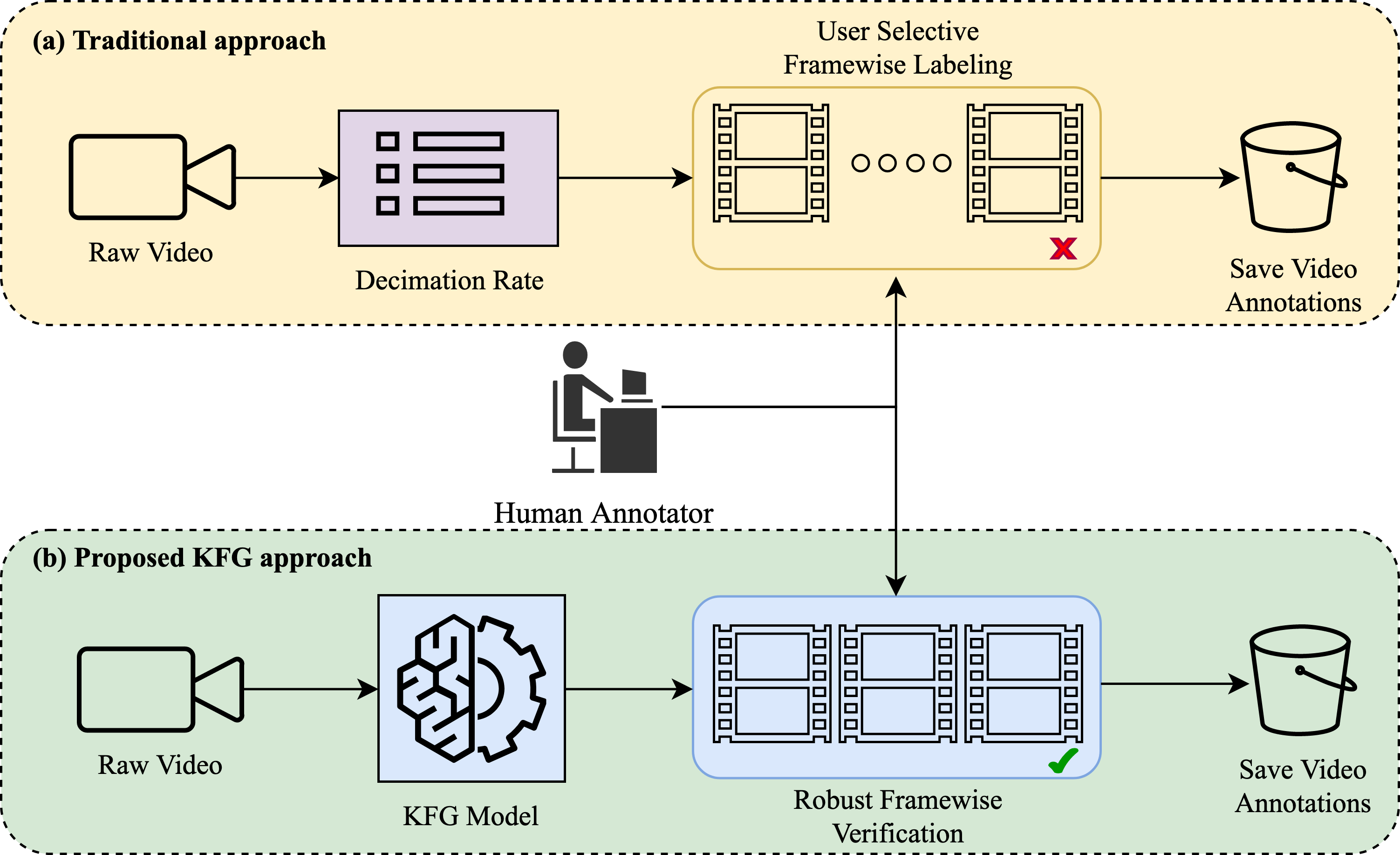}
  \caption{Comparison of key frame generation methods. (a) traditional approach: The raw video is down-sampled at a rate of 1/30 or 1/60 or frames per second (FPS), and video annotation is performed by the human annotator at decimated frames, followed by the saving annotations. (b) \textbf{KFG Solution}: The raw video is first processed by a key frame generation (KFG) machine learning (ML) model to generate robust key frames. Only the robust key frames are stored in the database (either in an S3 bucket or DynamoDB). Subsequently, the raw video, along with the generated robust key frames, are processed through annotation platforms, such as widely used SMGT, for human labeling. Later, the human annotations will be saved in the database for quality tasks, such estimation of annotated labels.} \label{block-diagram}
\end{figure}
Figure.\ref{block-diagram} shows the comparison between conventional retail video annotation approaches and our proposed machine learning-based method for identifying key frames that provide robust frame selection in retail environments. We named our demonstrated method the Key Frame Generation (KFG) ML model, specifically tailored for retail applications. We utilized object detection-based deep learning models to identify retail-relevant objects with annotations and select frames that require human annotator attention (such as verification to adjust bounding boxes) based on detected object probability. Specifically, we employed the YOLO (You Only Look Once) family methods \cite{yolov5}, such as YOLOv5x and YOLOv8x, to detect bounding boxes along with their class (customer, staff, product, shopping cart) and detection probability (ranging from 0 to 1). The YOLO models have multiple versions, including YOLOv5, YOLOv8, and the latest YOLONAS, each varying in object detection accuracy and computational requirements. For our retail experiments, we opted for YOLOv5x and YOLOv8x models that provide highly accurate predictions of in-store interactions.

Our demonstrated approach features three cases based on detection object probability in retail scenes:
\begin{itemize}
    \item High-confidence detections: We set a threshold for detection probability in the object detection models (YOLO) and selected frames that surpassed this detection threshold (example: 0.5 as threshold $p_{det}>th_1$). These frames don't need any human annotator involvement, and we use the annotations as they are. These typically represent clear customer-product interactions or well-defined retail scenarios.
    \item Medium-confidence detections: We set a threshold range where the detection object probability falls within an intermediate range (example: between 0.3 to 0.5 as $th_2 < p_{det} < th_1$). For these frames, a human annotator needs to validate/verify the annotations by editing/correcting the class and bounding box positions. These often represent more complex retail scenarios such as multiple customers interacting with the same product display or partially occluded merchandise.
    \item Low-confidence detections: Finally, we set a threshold range for frames that can be skipped because the object detection probability is low (example: less than 0.3 as $<th_2$). Annotations for these frames can be derived using interpolation methods (either linear or cubic-spline interpolation). These typically represent transitional moments in shopping behavior or periods with minimal activity.
\end{itemize}
To limit the number of frames requiring human verification to only the most critical key frames, the system administrator can select a larger threshold value during object detection (selecting only low-confidence frames for human annotation). This is particularly valuable for retailers who need to process large volumes of in-store footage efficiently. We validated the annotation accuracy by comparing the KFG-selected frames with ground truth labels (please see Section.\ref{metrics}) using the intersection-over-union (IOU) metric. For example, in a test video of retail store footage, we performed object detection with specific threshold values and recorded the detected frames along with their bounding boxes. We then compared the accuracy of these detected frames with a professionally annotated retail video dataset. More details about the number of detected frames and IOU metrics in retail scenarios can be found in Appendix.\ref{appendix_d}. A significant advantage of our KFG method for retail applications is that we can compare end-to-end annotation accuracy, since bounding boxes are available during the object detection step. This enables retailers to assess the quality of automated annotations and make informed decisions about when human intervention is necessary, optimizing both cost and annotation quality for critical retail analytics tasks.

\section{Experiments} \label{sec3}
To demonstrate our KFG method's applicability to retail environments, we selected commercial video annotation datasets that feature retail scenarios. In this paper, we focus on a commercial retail video dataset for validation of the KFG method and the open-access dataset MOT15\footnote{https://motchallenge.net/data/MOT15/} 
 for illustration purposes, adapting it to simulate retail contexts.
 
\subsection{Dataset} \label{dataset}
In this study, we utilize a dataset containing retail videos that were labeled using human annotators within a standardized video annotation framework. The dataset comprises 935 videos captured from various in-store surveillance systems and cameras with different resolutions, frame sizes, and frame rates. The footage includes diverse retail environments such as supermarkets, apparel stores, electronics retailers, and convenience shops, providing a comprehensive view of different shopping scenarios and customer behaviors. The annotations were performed by two or more human annotators for each video, with a human auditor verifying the accuracy of the video annotations to ensure high-quality ground truth data. Table~\ref{dataset_1} shows the distribution of video resolutions in our retail dataset. The frame rate varies from 2 to 33 frames per second (fps), though approximately 80\% of the retail videos have a frame rate around 14 fps, which is typical for retail surveillance systems that balance detail with storage requirements.
\begin{table}[!ht]
\small
\caption{Resolution distribution of the videos in the video dataset.}
\centering
\begin{tabular}{cc}
\hline
\textbf{Frame resolution} & \textbf{Number of videos} \\ \hline
360x640          & 88       \\ 
480x848          & 3        \\ 
720x1280         & 228      \\ 
1080x1920        & 616      \\ \hline
\end{tabular}
\label{dataset_1}
\end{table}
\begin{figure}[!ht]
  \centering
  \includegraphics[width=0.45\linewidth]{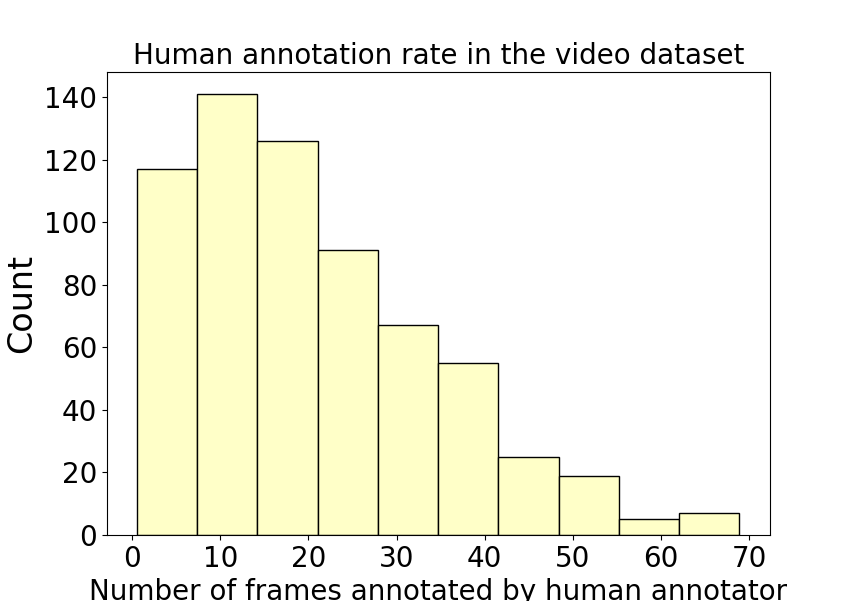}
  \caption{Annotation of the ground truth key frames in the Video dataset (includes person, animal and vehicle classes).} \label{annotation-diagram}
\end{figure}
The distribution of human annotations in the video dataset per class type shows approximately 70\% for customers (person class), 12.3\% for pets (animal class in pet stores), and 78.5\% for shopping carts and vehicles in parking areas (vehicle class). In this retail dataset, the number of frames in raw videos ranges from 82 to 1,928 frames, with an average of 635 frames per video. From the annotation data focusing on customers (person class), the number of annotated frames ranges from 4 to 749, with an average of 118 frames annotated per video. This substantial annotation effort is expensive (see Appendix~\ref{appendix_a} for cost details) and can be significantly reduced using our demonstrated KFG approach, resulting in considerable cost savings for retailers.

In addition to our commercial retail dataset, we adapted the MOT15 dataset, an open-access resource, to evaluate our method by selecting sequences that resemble retail environments and customer movement patterns. Figure~\ref{annotation-diagram} shows the annotation distribution across videos with customers, pets, and vehicles/carts as class types. The x-axis represents the annotation rate (number of key-frames/total number of frames), while the y-axis shows the count of videos with each particular annotation rate. This visualization reveals that on average, 17.6\% (median value) of frames are annotated as key frames in the retail video dataset for the customer class. For videos specifically, there are 352 in the dataset, where the ground truth annotation rate is 15.54\% (median value) across customer, pet, and shopping cart/vehicle classes. This dataset presents an ideal test case for our KFG method, as retail environments combine challenges of multiple humans exist in the frame/video, identifying interactions with static objects (products), and analyzing complex behaviors (shopping patterns) - all scenarios where efficient key frame selection is critical for cost-effective annotation.

\subsection{Metrics} \label{metrics}
\begin{figure}[!ht]
  \centering
  \includegraphics[width=0.6\linewidth]{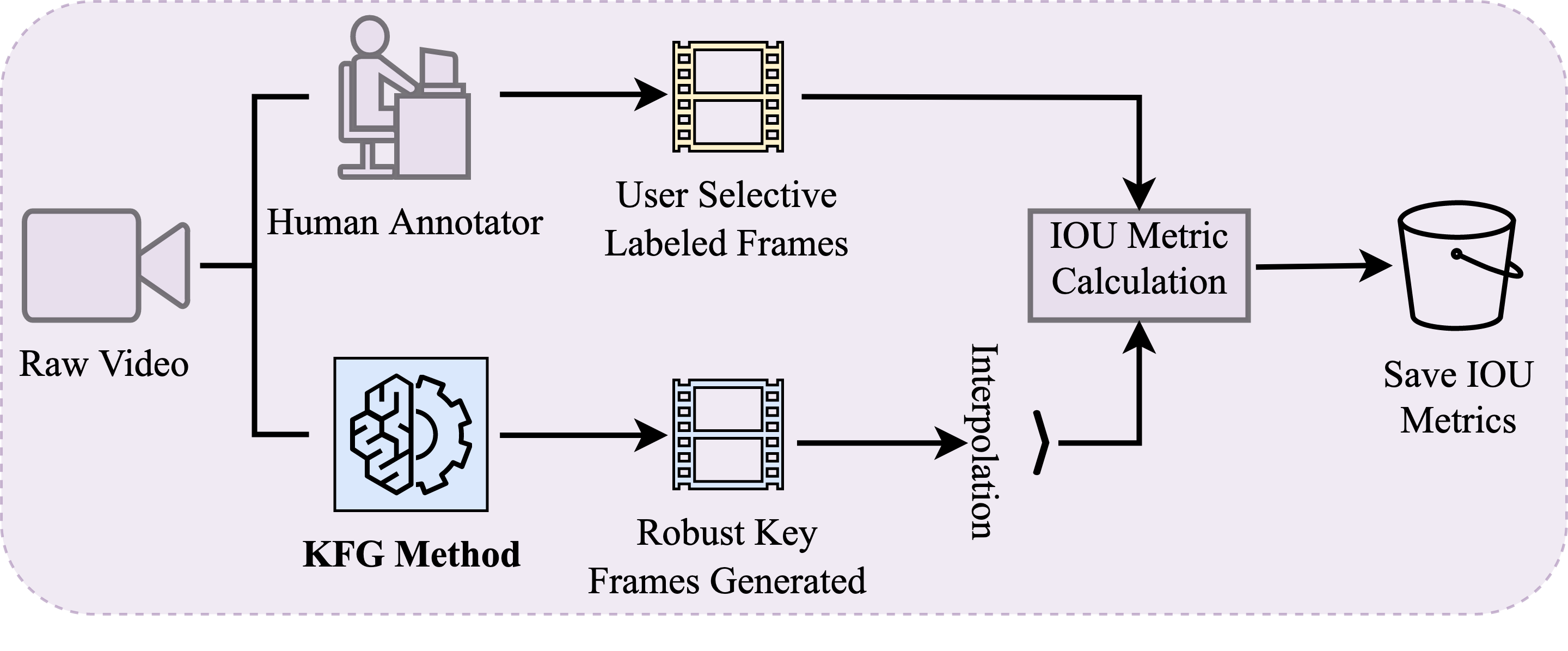}
  \caption{Intersection-over-Union (IOU) metrics calculation using the human annotator selective labeled frames and the KFG method predicted annotations, followed by the interpolation in the video annotation dataset. Finally, the results are saved to a common location.} \label{eval-diagram}
\end{figure}
In this paper, we use the mean intersection-over-union (IOU) metric \cite{gawande2020deep} on complete retail video annotations as the standard qualitative measure to evaluate the performance of different key-frame identification methods in retail environments. This metric is particularly suitable for retail applications as it assesses how accurately our automated system identifies customer-product interactions compared to human annotator labeling. The IOU metric provides a rigorous evaluation of bounding box accuracy, which is critical in retail analytics where precise identification of customer interactions with products influences merchandising decisions and store layout optimization. Higher IOU values indicate better alignment between our automated annotations and human-verified ground truth, confirming that our key-frame selection correctly captures relevant retail behaviors. Figure~\ref{eval-diagram} shows the evaluation metrics calculation block diagram for our demonstrated KFG approach using human annotator video annotations in the commercial retail video dataset. Our evaluation pipeline first processes raw retail footage through both our KFG system and traditional human annotation workflows. We then compare the resulting annotations using the IOU metric to quantify the accuracy of our automated approach in detecting customer movements, product interactions, and shopping behaviors typical in retail environments. This evaluation framework allows retailers to understand the trade-offs between annotation cost, speed, and accuracy when implementing our KFG system, providing them with quantifiable evidence of the benefits our approach offers for large-scale retail video analysis and annotation tasks.

\subsection{Results} \label{results_discuss}
In this section, we compare different key-frame generation methods and propose our approach for real-time key-frame detection in retail environments, where human annotators verify, adjust, and annotate only frames with low confidence scores. This is particularly valuable in retail settings where annotation budgets are constrained but accuracy in capturing customer-product interactions remains critical. We compare three approaches: traditional methods without deep learning, deep learning with frame encoding followed by K-means clustering, and our KFG method which employs object-detection for key-frame identification in retail videos.

\subsubsection{Traditional method: FFMPEG Addon} \label{trad_methods}
\begin{figure}[!t]
    \centering
    \subfloat{{\includegraphics[width=0.43\linewidth, height=.4\linewidth]{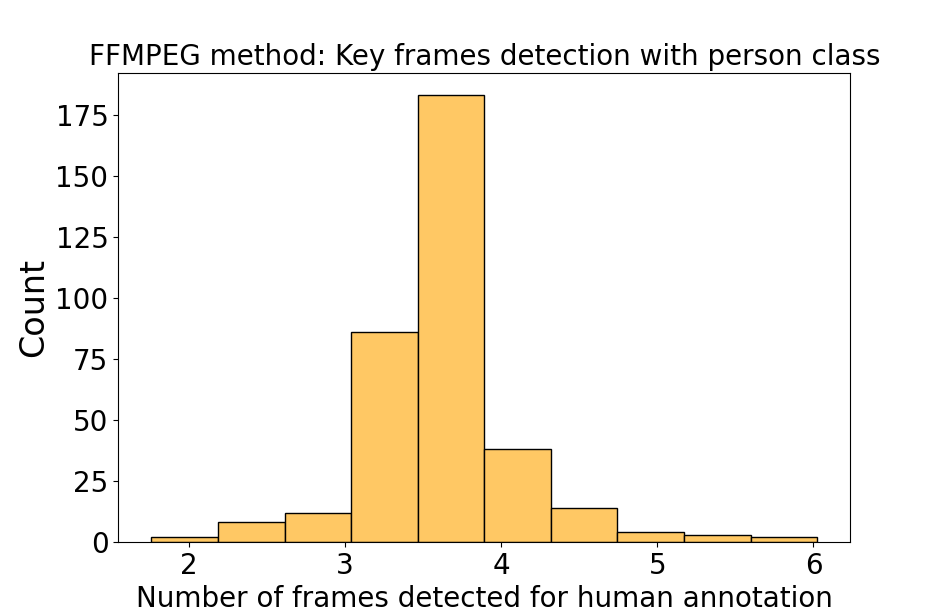}}}%
    \qquad 
    \subfloat{{\includegraphics[width=0.43\linewidth, height=.4\linewidth]{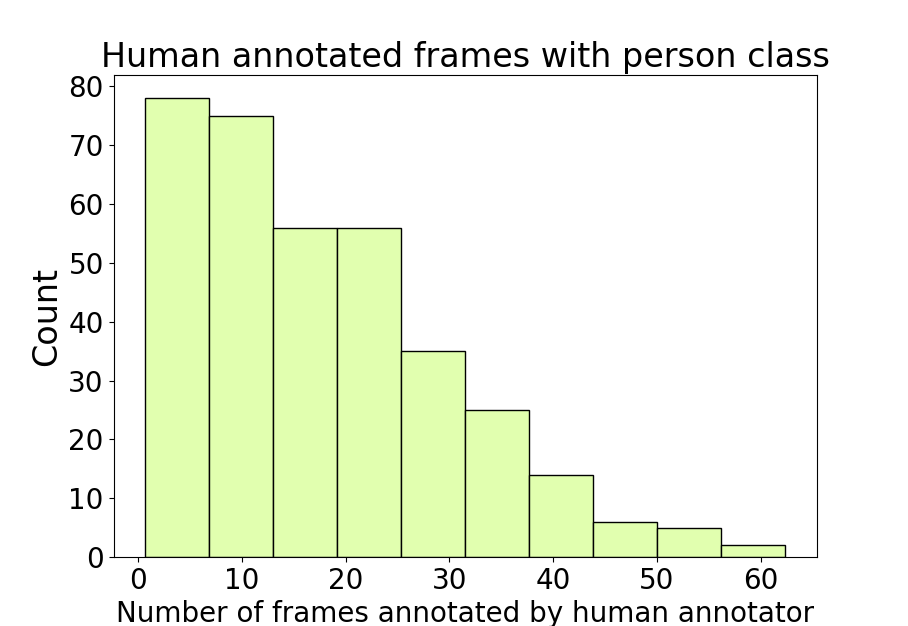}}}%
    \caption{Illustration of key-frames annotation rate using FFMPEG addon and compare with the human annotations where videos contain person class. Leftside: Detected key frames using FFMPEG method on the video dataset where videos contain only person class. Rightside: Illustration of human annotation rate in the video dataset contains only the person class.}%
    \label{ffmpeg}
    \vspace{-0.6cm}
\end{figure}

Traditional methods explained in Section~\ref{td_methods} can identify key frames but lack the ability to generate bounding boxes necessary for high-quality retail annotation. Figure~\ref{ffmpeg} shows the results of FFMPEG addon detecting intra-frames (I-frames) in retail videos where each video contains at least one customer present (and corresponding human frame-wise annotations with customer class labels). We specifically selected videos containing at least one customer to avoid analyzing empty store footage where no annotations would be required. While the annotation rate appears improved using FFMPEG addon to detect key frames in retail videos, the critical retail-specific annotation quality using bounding boxes for customer-product interactions is missing. This limitation significantly reduces the utility of this approach for retail analytics that depend on precise interaction identification.

Quantitatively, human frame-wise annotation in our retail video dataset contains average annotation/key-frames of 15.54\% (median value) for the customer class, while FFMPEG addon reduces the annotation cost by identifying key frames at a rate of 3.64\% (median value). This suggests potential annotation cost savings of approximately 4.27 times. However, without bounding box capabilities, this approach is unable to deliver the specific annotation quality required for sophisticated retail analytics applications such as customer journey mapping, product interaction analysis, or theft prevention.

\subsubsection{Deep learning method: Video encoding with K-means clustering} \label{encoding}

Deep learning methods explained in Section~\ref{dl_methods} can identify distinctive key frames in retail videos but, like traditional methods, lack bounding box capabilities essential for comprehensive retail annotation quality. To evaluate this approach in retail contexts, we performed ResNet-50 based encoding of retail video frames and applied normalization (with zero mean and unit variance) on the latent dimension before applying PCA on the encoded representations. We then implemented unsupervised K-means clustering on the normalized encoded embeddings to identify distinct shopping scenarios and customer behaviors within retail footage. From our experiments with in-store videos, selecting the optimal value for "K" in K-means clustering proved critical for providing robust key-frames that captured the diversity of retail interactions. To determine the ideal K-value, we employed the Elbow method and identified a "K" value of 10 as optimal for most retail test videos, representing different phases of customer shopping behavior (e.g., browsing, product examination, checkout).

We observed that the centroids of these clusters effectively served as key frames representing distinct shopping activities within retail videos. This approach dramatically reduced the number of frames requiring annotation to a fixed value of 10 (the K-value in K-means clustering), which represents just 1.98\% (median value) of total frames in typical retail videos. However, this method does not facilitate comparison of video annotation quality since it doesn't generate bounding boxes for shoppers, staff, or products - a critical limitation for retail applications that need to identify specific interactions. Despite this limitation, the approach demonstrates potential annotation cost savings of approximately 7.85 times for retailers. This represents significant efficiency gains for initial video summarization, but would require supplementary techniques to provide the detailed bounding box annotations necessary for comprehensive retail analytics. The clustering approach does offer valuable insights for retailers by automatically identifying distinct shopping behavior patterns, which could complement our KFG method in a hybrid approach for more sophisticated retail video analysis. This combination could potentially identify both key behavioral moments (through clustering) and specific object interactions (through our object detection approach).

\begin{table*}[!ht]
\caption{\label{anno_res1} KFG based Object detection to show the number of videos and median frames detected (\%) along with annotation on the video dataset using YOLOv5x and YOLOv8x models.}
\tiny
\centering
\begin{tabular}{l|llllllll|l}
\hline
\multicolumn{9}{c|}{\textbf{Detection threshold ($th_1$) total videos with only person class: 330}}  &   \\ \hline
\begin{tabular}[c]{@{}l@{}}\textbf{Metrics} \end{tabular}  & \multicolumn{1}{l}{0.2}   & \multicolumn{1}{l}{0.3}   & \multicolumn{1}{l}{0.4}   & \multicolumn{1}{l}{0.5}   & \multicolumn{1}{l}{0.6}   & \multicolumn{1}{l}{0.7}   & \multicolumn{1}{l}{0.8}  & 0.9  & human annotations \\ \hline 
\multicolumn{10}{c}{YOLOv5x}   \\ \hline
\textbf{Median frames detected: \%}  & \multicolumn{1}{l}{24.17} & \multicolumn{1}{l}{22.50} & \multicolumn{1}{l}{19.90} & \multicolumn{1}{l}{17.95} & \multicolumn{1}{l}{15.49} & \multicolumn{1}{l}{14.56}  & \multicolumn{1}{l}{11.57} & 7.07 & 16.01   \\ 
\textbf{\#Videos }                                                          & \multicolumn{1}{l}{330}   & \multicolumn{1}{l}{327}   & \multicolumn{1}{l}{321}   & \multicolumn{1}{l}{316}   & \multicolumn{1}{l}{309}   & \multicolumn{1}{l}{300}   & \multicolumn{1}{l}{284}  & 214  & 330     \\ \hline
\multicolumn{10}{c}{YOLOv8x}   \\ \hline
\textbf{Median frames detected: \%}                                                                       & \multicolumn{1}{l}{22.65} & \multicolumn{1}{l}{21.52} & \multicolumn{1}{l}{20.73} & \multicolumn{1}{l}{18.98} & \multicolumn{1}{l}{17.53} & \multicolumn{1}{l}{16.86}  & \multicolumn{1}{l}{13.90} & 8.49 & 16.01   \\ 
\textbf{\#Videos}                                                          & \multicolumn{1}{l}{330}   & \multicolumn{1}{l}{326}   & \multicolumn{1}{l}{320}   & \multicolumn{1}{l}{314}   & \multicolumn{1}{l}{309}   & \multicolumn{1}{l}{299}   & \multicolumn{1}{l}{283}  & 220  & 330     \\ \hline
\end{tabular}
\end{table*}

\subsubsection{KFG model: object detection based key-frame detection} \label{yolo_method}
\begin{table*}[!ht]
\caption{\label{final-result} Cost saving (automatic annotations of the videos and frames \%) and accuracy of various methods to detect the key-frames in the video annotation performed on the video dataset where each video contains atleast one person.}
\tiny
\centering
\begin{tabular}{l|l|l|l|l|l}
\hline
\textbf{Technique}   & \begin{tabular}[c]{@{}l@{}} KFG annotated \\ videos (\%)\end{tabular} & \begin{tabular}[c]{@{}l@{}}KFG annotated \\ frames (\%)\end{tabular} & \begin{tabular}[c]{@{}l@{}}Videos  need  \\ human annotations (\%)\end{tabular} & \begin{tabular}[c]{@{}l@{}}Human annotated \\ frames (\%)\end{tabular} & \begin{tabular}[c]{@{}l@{}}Automated frames \\ IOU metric \end{tabular}   \\ \hline
FFMPEG addon & NA    & NA  & 100    & 3.64    & NA   \\ \hline
\begin{tabular}[c]{@{}l@{}}Video encoding \\ with clustering\end{tabular}   & NA   & NA  & 100   & 1.98    & NA   \\ \hline
\begin{tabular}[c]{@{}l@{}}KFG with \\ YOLOv5x ($th_1= 0.5$)\end{tabular}   & 95.75  & 17.95    & 4.25   & NA  & 0.51 \\ \hline
\begin{tabular}[c]{@{}l@{}}KFG with \\ YOLOv8x ($th_1 =  0.5$)\end{tabular} & 95.15  & 18.98   & 4.85   & NA   & 0.48 \\ \hline
\end{tabular}
\end{table*}
KFG method: In our proposed deep learning-based key frame detection method for retail environments, we leverage state-of-the-art object detection (YOLOv5 or YOLOv8 models) on video frames followed by intelligent frame selection based on detection probability thresholds. This approach is specifically optimized for retail scenarios where precise identification of customers, staff, and product interactions is crucial. We computed IOU calculations with varying thresholds on our comprehensive retail video dataset, limiting analysis to videos containing at least one customer along with corresponding human-annotated ground truth. Table~\ref{anno_res1} shows the detected frames at various thresholds from 0.2 to 0.9, with detection frame rates ranging from 24.17\% to 7.07\% in retail footage. When applying a threshold of 0.8 to in-store video, 11.57\% of frames are automatically detected with high confidence, requiring no human verification. This annotation rate of 0\% (for human intervention) is substantially lower than the traditional human annotation rate of 16.01\%, representing significant efficiency gains for retail annotation workflows. At this high probability threshold, YOLOv5x model detections are highly accurate, making them immediately suitable for downstream retail analytics tasks such as shopper journey analysis and product interaction detection.

This method substantially assists human annotators by providing high-quality pre-annotations for retail footage, reducing the average handling time for video annotation tasks - a critical efficiency metric for retailers managing large volumes of surveillance footage. It's important to note that at the detection probability threshold of 0.8, the system successfully processed 284 out of 330 (86\%) retail videos containing customer interactions, with the remaining videos (14\%) requiring traditional manual annotation due to challenging lighting conditions or unusual customer behaviors. Similar results were achieved with the YOLOv8x model. At the detection threshold of 0.5, which balances precision with recall, only 14 and 16 videos (out of 330) required human labeling when using YOLOv5 and YOLOv8 models, respectively. This translates to potential annotation cost savings of approximately 20 times for retailers, even before considering IOU metrics for quality assurance.

To comprehensively evaluate annotation quality in retail contexts, we followed the method described in Section~\ref{metrics}, using linear interpolation for frames where the detected probability fell below the threshold or where detection was missing entirely - common in situations like product occlusion or customer crowding. Table~\ref{final-result} compares various methods for detecting key frames in retail video datasets, evaluating both cost savings and annotation quality using IOU metrics. The YOLOv5x-based approach provides the most accurate key-frame generation with higher quality bounding box detection in retail scenarios, with lower confidence frames efficiently annotated using interpolation.

Our proposed KFG method delivered automatic annotations for approximately 95\% of retail videos (316 out of 330) with an IOU metric exceeding 0.5, completely eliminating the need for human verification in these cases. Only 5\% of videos (14 out of 330) required traditional annotation, significantly reducing Average Annotation Time (AAT) and annotation costs for retail operations. Appendix~\ref{appendix_d} details the relationship between detection thresholds and calculated IOU values for test retail videos. Our experiments revealed that higher thresholds provide more accurate annotations but reduce the number of detected frames, potentially decreasing the IOU metric due to sparse detection. Therefore, finding the optimal threshold balancing detection frequency against object accuracy is essential for retail applications, where different scenarios (e.g., busy weekend shopping vs. quieter weekday periods) might benefit from different threshold configurations.

This adaptive threshold approach makes our KFG method particularly valuable for retailers, who can tune the system to prioritize either annotation cost savings or detection precision based on their specific business needs and video analytics requirements.

\begin{table*}[!ht]
\caption{\label{iou_yolos} KFG based key frame identification on the video dataset using YOLOv5x and YOLOv8x models with detection threshold as 0.5.}
\small
\centering
\begin{tabular}{l|llllllll}
\hline
& \multicolumn{8}{c}{\textbf{\#videos with IOU (>$th_{iou}$) with total videos: 330}}   \\ \hline
\begin{tabular}[c]{@{}l@{}}\textbf{Method}\\ ($th_{iou}$) \end{tabular} & \multicolumn{1}{l}{0.2}   & \multicolumn{1}{l}{0.3}   & \multicolumn{1}{l}{0.4}   & \multicolumn{1}{l}{0.5}   & \multicolumn{1}{l}{0.6}   & \multicolumn{1}{l}{0.7}   & \multicolumn{1}{l}{0.8}  & 0.9   \\ \hline
\textbf{Yolov5x}    & \multicolumn{1}{l}{255} & \multicolumn{1}{l}{226} & \multicolumn{1}{l}{191} & \multicolumn{1}{l}{155} & \multicolumn{1}{l}{115} & \multicolumn{1}{l}{77}  & \multicolumn{1}{l}{34} & 8  \\ 
\textbf{Yolov8x}   & \multicolumn{1}{l}{270} & \multicolumn{1}{l}{229} & \multicolumn{1}{l}{191} & \multicolumn{1}{l}{156} & \multicolumn{1}{l}{116} & \multicolumn{1}{l}{75} & \multicolumn{1}{l}{30} & 6   \\ 
\hline
\end{tabular}
\end{table*}

\begin{figure}[!ht]
    \centering
    \subfloat{{\includegraphics[width=0.45\linewidth, height=.4\linewidth]{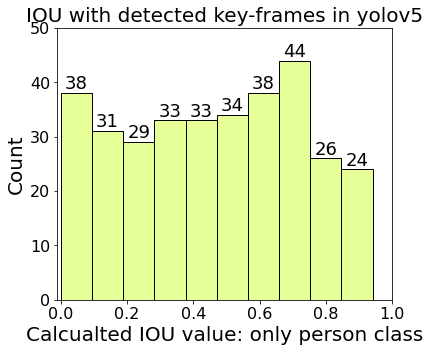}}}%
    \qquad
    \subfloat{{\includegraphics[width=0.45\linewidth, height=.4\linewidth]{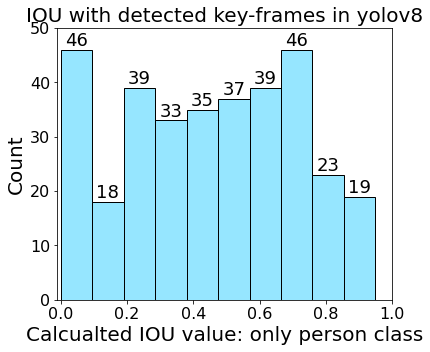}}}%
    \caption{Illustration of videos count vs. calculated IOU when using the key-frames including interpolation in the KFG method and compare with the human annotations (only person class is selected).Leftside: YOLOv5: Number of videos vs calculated IOU on the video dataset where the videos contain only person class., Rightside: YOLOv8: Number of videos vs calculated IOU on the video dataset where the videos contain only person class. }%
    \label{iou_plots}%
\end{figure}

Figure~\ref{iou_plots} illustrates the distribution of videos with their corresponding IOU scores calculated using our KFG method in retail environments, compared against frame-wise human annotations from our retail video annotation dataset. This visualization clearly demonstrates how our automated approach performs across diverse retail scenarios including different store layouts, lighting conditions, and customer densities. Table~\ref{iou_yolos} provides a detailed breakdown of video counts where the calculated IOU value exceeds specified thresholds when compared with human annotations for both YOLOv5 and YOLOv8 models in retail contexts. The results confirm that KFG-based YOLOv5x and YOLOv8x models effectively detect and annotate customer movements, staff activities, and product interactions in retail video footage with minimal human intervention.

This capability enables retail annotation teams to focus exclusively on frames where detection probability falls below a certain threshold (in this case 0.5), requiring adjustment only for low-confidence detections such as partially occluded customers or complex multi-person product interactions. The data shows that approximately 46.96\% of retail videos (155 out of 330) can be automatically annotated with an IOU exceeding 0.5 using our KFG model with YOLOv5-based detection. This translates to annotation cost savings of at least 2× times while maintaining the high-quality annotations essential for accurate retail analytics. Our demonstrated KFG-based key-frame detection method significantly reduces video annotation costs for retailers, who typically process thousands of hours of footage weekly across multiple store locations. The approach can be applied across various retail-specific dataset creation tasks and downstream analytics, including customer journey mapping, product interaction analysis, theft prevention, and store layout optimization.

Additionally, this method serves as an effective co-pilot for retail video annotation teams, assisting human annotators during real-time annotation sessions by pre-filling likely detections and significantly speeding the workflow. For retailers with specific product recognition needs, fine-tuning the models is recommended to achieve higher detection probability for store-specific items or unique display fixtures (though such customization is beyond the scope of this current research). By implementing our KFG approach, retailers can dramatically improve the efficiency of their video annotation processes while maintaining or even enhancing the quality of resulting analytics, ultimately supporting more data-driven decision making in merchandising, store design, and customer experience optimization.

\section{Conclusions and Future work}
In conclusion, this research presents a deep learning-based Key Frame Generation (KFG) method specifically designed to enhance retail video labeling efficiency. The KFG approach accurately detects key frames and generates high-quality annotations of customer-product interactions, showcasing promising results for retail environments. Compared to traditional methods, KFG provides robust frame selection, improved accuracy, greater efficiency, and significant annotation cost savings - with less than 5\% of frames requiring human annotation or verification. This valuable contribution to retail video analysis paves the way for more efficient and accurate video annotation processes that directly impact retail operations and analytics. 
Future research directions for our retail video analysis using the KFG methodology include: enhancing product-customer interaction detection to provide deeper shopping behavior insights; integrating multi-modal retail data to create comprehensive shopper journey maps; incorporating privacy-preserving capabilities to address growing privacy concerns; and developing adaptive real-time annotation systems for immediate insights and responsive retail experiences. By continuing to refine the KFG method for retail, we aim to deliver valuable tools that transform video data into actionable retail intelligence while reducing the resources required for video annotation.

\bibliographystyle{unsrt}
\bibliography{alexa_automated_annotation}

\section*{Appendix A: Annotation cost calculation} \label{appendix_a}
This section shows the annotation cost per object per frames in the video annotation using the amazon web services (AWS) sagemaker ground truth (SMGT) platform which is one of the video annotation platform. For example, let's consider a video contains a human walking in a street and has a frame rate of 30 FPS. Video duration is 1 minute (60-seconds), results total frames as 1800. Bounding box annotation cost is $0.036\$$ per object, which will result a total cost of 64.8\$. In addition, this cost will be increased linearly if more annotators are required for the high-quality labels. If the DA selects the annotation at FPS, then the total cost is 2.16\$. However, our demonstrated approach annotation cost is only \textbf{0.72\$} with selected key frames as 20. Additional details about the annotation cost can be found in this web page\footnote{https://aws.amazon.com/sagemaker/data-labeling/pricing/}.

\section*{Appendix B: Key-frames detection using YOLO} \label{appendix_d}
In this section, we show the settings of the YOLOv5x (YOLO version 5 and xlarge model) model where the image is resized to 640$\times$640, RGB image, probability threshold to detect an object is set to a value that can be tuned based on the requirement and only selected the detection class type as person in the video dataset. For example consider a test video from the video dataset with image resolution is 1080$\times$1920, with fps of 14.95. We varied the threshold from 0.5 to 0.8 and found the detected frames and corresponding bounding boxes. Table.\ref{iou_tab1} indicates for a given test video, the detected key-frames and the corresponding IOU metric when compared the ground truth. Note that selecting the higher threshold provides the accurate annotations but the detected number of frames reduces in a given video for a class (person class) that decreases the IOU metric value due to sparse frames detection.

\begin{table}[!ht]
\caption{Trade of between the detected frames and IOU metric on a test video in the video dataset.}\label{iou_tab1}
\small
\centering
\begin{tabular}{l|llll|l}
\hline
 & \multicolumn{4}{c|}{Total frames in test video: 335}      &   Ground truth  \\ \hline
\textbf{Metrics} & \multicolumn{1}{c}{0.50}  & \multicolumn{1}{c}{0.60}  & \multicolumn{1}{c}{0.70}  & 0.80 & NA \\ \hline
\begin{tabular}[c]{@{}l@{}}Detected\\ frames (\#) \end{tabular} & \multicolumn{1}{c}{58} & \multicolumn{1}{c}{45} & \multicolumn{1}{c}{27} & 4 & 37  \\ \hline
\textbf{Mean IOU}  & \multicolumn{1}{c}{0.59}  & \multicolumn{1}{c}{0.58}  & \multicolumn{1}{c}{0.59}  & 0.29 & NA  \\ \hline
\end{tabular}
\end{table}

\end{document}